\documentclass[conference]{IEEEtran}
\usepackage[latin9]{inputenc}
\usepackage{graphicx}
\usepackage{stfloats}
\usepackage{bm}
\usepackage{amssymb}
\usepackage{amsmath}
\usepackage{color}
\usepackage{balance}

\renewcommand{\r}   {{\bf r}}

\newcommand{\ci}    {{\bf c}}

\newcommand{\hr}    {\widehat{r}}

\newcommand{\X}     {\bf X}
\newcommand{\R}     {\bf R}
\newcommand{\hR}    {\widehat{\bf R}}

\newcommand{\ru}    {\rule{0mm}{3.4mm}}

\newcommand{\be}    {\begin{equation}}
\newcommand{\ee}    {\end{equation}}

\begin{document}

\title{Recasting Residual-based Local Descriptors \\ as Convolutional Neural Networks: \\ an Application to Image Forgery Detection}

\author{
       \IEEEauthorblockN{Davide Cozzolino, Giovanni Poggi, Luisa Verdoliva\\}
       \IEEEauthorblockA{DIETI, University Federico II of Naples, Italy\\
       Email: \{davide.cozzolino, poggi, verdoliv\}@unina.it}
}
\maketitle

\begin{abstract}
Local descriptors based on the image noise residual have proven extremely effective for a number of forensic applications,
like forgery detection and localization.
Nonetheless, motivated by promising results in computer vision, the focus of the research community is now shifting on deep learning.
In this paper we show that a class of residual-based descriptors
can be actually regarded as a simple constrained convolutional neural network (CNN).
Then, by relaxing the constraints, and fine-tuning the net on a relatively small training set,
we obtain a significant performance improvement with respect to the conventional detector.
\end{abstract}


\IEEEpeerreviewmaketitle

\section{Introduction}
\label{sec:intro}

Images and videos represent by now the dominant source of traffic on the Internet and the bulk of data stored on social media.
Nowadays, however, such data may be easily manipulated by malicious attackers
to convey some twisted and potentially dangerous messages.
Interested areas include politics, journalism, judiciary, even the scientific world.
For this reason, in the last few years there has been intense and ever growing activity on multimedia forensics, aiming at developing methods to detect, localize, and classify possible image
manipulations.

Typical attacks consist in adding or deleting objects,
using material taken from the same image (copy-move) or from other sources (splicing).
As a consequence, many researchers have focused on detecting near-duplicates in the image or across a repository of images.
More fundamentally, one may be interested in establishing
whether the image under analysis is pristine or else has been subjected to some post-processing,
since any form of manipulation may raise suspects and suggest deeper inquiry.
In fact, most of the times,
copy-moves and splicing are accompanied by various forms of elaboration aimed at removing the most
obvious traces of editing.
These include, for example, resizing, rotation, linear and non-linear filtering,
contrast enhancement, histogram equalization and, eventually, re-compression.

A number of papers have been proposed to detect one or the other of such elaborations
\cite{Popescu2005, Kirchner2008, Kirchner2010, Huang2010, Stamm2010}.
These methods, however, are sensitive to just some specific manipulations.
A more appealing line of research is to detect {\em all} possible manipulations,
an approach that has been followed in several papers \cite{Cao2012, Verdoliva2014, Fan2015, Li2016}.
Notably, in the 2013 IEEE Image Forensics Challenge,
the most effective techniques for both image forgery detection \cite{Cozzolino2014a} and localization \cite{Cozzolino2014b}
used this approach, relying on powerful residual-based local descriptors.
These features,
such as SPAM (subtractive pixel adjacency matrix) \cite{Pevny2010} or SRM (spatial rich models) \cite{Fridrich2012},
inspired to previous work in steganalysis, are extracted from the so-called residual image.
In fact, the noise residual, extracted through some high-pass filtering of the image,
contains a wealth of information on the in-camera and out-camera processes involved in the image formation.
Such subtle traces, hardly visible without enhancement,
may reveal anomalies due to object insertion \cite{Cozzolino2015b, Cozzolino2016}
or can detect different types of image editing operations \cite{Verdoliva2014, Li2016, Boroumand2017}.

Very recently,
inspired by impressive results in the closely related fields of computer vision and pattern recognition \cite{Krizhevsky2012},
the multimedia forensics community began focusing on the use of deep learning \cite{Bengio2013, LeCun2015},
especially convolutional neural networks (CNN) \cite{Zeiler2014}.
Taking advantage of the lesson learnt from SPAM/SRM features,
constrained CNN architectures have been proposed both for steganalysis \cite{Qian2015} and manipulation detection \cite{Bayar2016},
where the first convolutional layer is forced to perform a high-pass filtering.

In this paper we show that there is no real contraposition between residual-based features and CNNs.
Indeed, these local features can be computed through a CNN with architecture and parameters selected so as to guarantee a perfect equivalence.
Once established this result,
we go beyond emulation, removing constraints on parameters, and fine-tuning the net to further improve performance.
Since the resulting network has a lightweight structure,
fine-tuning can be carried out by means of a limited training set, limiting computation time and memory usage.
A significant performance gain with respect to the conventional feature is observed, especially in the most challenging situations.

In the following we describe in more detail the residual-based local features (Section 2),
recast them as a constrained CNN, to be further trained after removing constraints (Section 3),
show experimental results (Section 4),
and draw conclusions (Section 5).

\section{Residual-based local descriptors}

Establishing which type of processing an image has undergone
calls for the ability to detect the subtle traces left by these operations,
typically in the form of recurrent micropatterns.
This problem has close ties with steganalysis, where weak messages hidden in the data are sought,
so it is no surprise that the same tools, residual-based local descriptors,
prove successful in both cases.
To associate a residual-based feature to an image, or an image block, the following processing chain
has been successfully used in steganalysis \cite{Pevny2010, Fridrich2012}:
\begin{enumerate}
	\item   extraction of noise residuals
	\item   scalar quantization
	\item   computation of co-occurrences
	\item   computation of histogram
\end{enumerate}

In the following, we describe in some more depth all these steps,
taking a specific model out of the 39 proposed in \cite{Fridrich2012} as running example.

\vspace{3mm} \noindent
{\bf Extraction of noise residual.}
The goal is to extract image details, in the high-frequency part of the image,
which enables the analysis of expressive micropatterns.
As the name suggests, this step can be implemented by resorting to a high-pass filter.
In \cite{Fridrich2012} a number of different high-pass filters have been considered,
both linear and nonlinear, with various supports.
Here, as an example,
we focus on a single 4-tap mono-dimensional linear filter, with coefficients
${\bf w} = [1, -3, 3, -1]$.
The filter extracts image details along one direction, but is applied also on the image transpose (assuming vertical/horizontal invariance) to augment the available data.
Choosing a single filter rather than considering all models proposed in \cite{Fridrich2012}
is motivated not only by the reduced complexity
but also by the very good performance observed in the context of image forgery detection \cite{Cozzolino2014a, Verdoliva2014, Li2016}.



\vspace{3mm} \noindent
{\bf Scalar quantization.}
Residuals are conceptually real-valued quantities or, in any case,
high-resolution integers, so they must be quantized to reduce cardinality and allow easy processing.
In \cite{Fridrich2012} a uniform quantization is used with an odd number of levels (to ensure that 0 is among the possible outputs).
Therefore, the only parameters to set are the number of quantization levels, $L$, and the quantization step $\Delta$.
In our example we set $L=3$ and $\Delta=4.5$.

\vspace{3mm} \noindent
{\bf Co-occurrences.}
The computation of co-occurrences is the core step of the procedure.
In fact, this is a low-complexity means for taking into account high-order dependencies among residuals
and hence gather information on recurrent micropatterns.
Following \cite{Fridrich2012} we compute co-occurrences on $N=4$ pixels in a row,
both along and across the filter direction.
With these values, two co-occurrence matrices with $3^4=81$ entries are obtained.
Of course, the image or block under analysis must be large enough to obtain meaningful estimates.
All the co-occurrence $N$-dimensional bins are eventually coded as integers.

\vspace{3mm} \noindent
{\bf Feature formation.}
Counting co-occurrences one obtains the final feature vector describing the image.
Neglecting symmetries, our final feature has length equal to 162.
The final classification phase is performed by a linear SVM classifier.

\section{Recasting local features as CNN}

We will now show that local residual-based features can be extracted by means of a convolutional neural network.
Establishing this equivalence leaves us with a CNN architecture and a set of parameters that are already known to provide an excellent performance for the problem of interest.
Then, given this good starting point,
we can move a step forward and fine-tune the network through a sensible training phase with labeled data.
Note that in this way we will carry out a joint optimization of both the feature extraction process and classification.
In the following we will first move from local features to a Bag-of-Words (BoW) paradigm,
and then proceed to the implementation by means of Convolutional Neural Networks.

\begin{figure}[t!]
    \centering
    \includegraphics[width=1.0\linewidth]{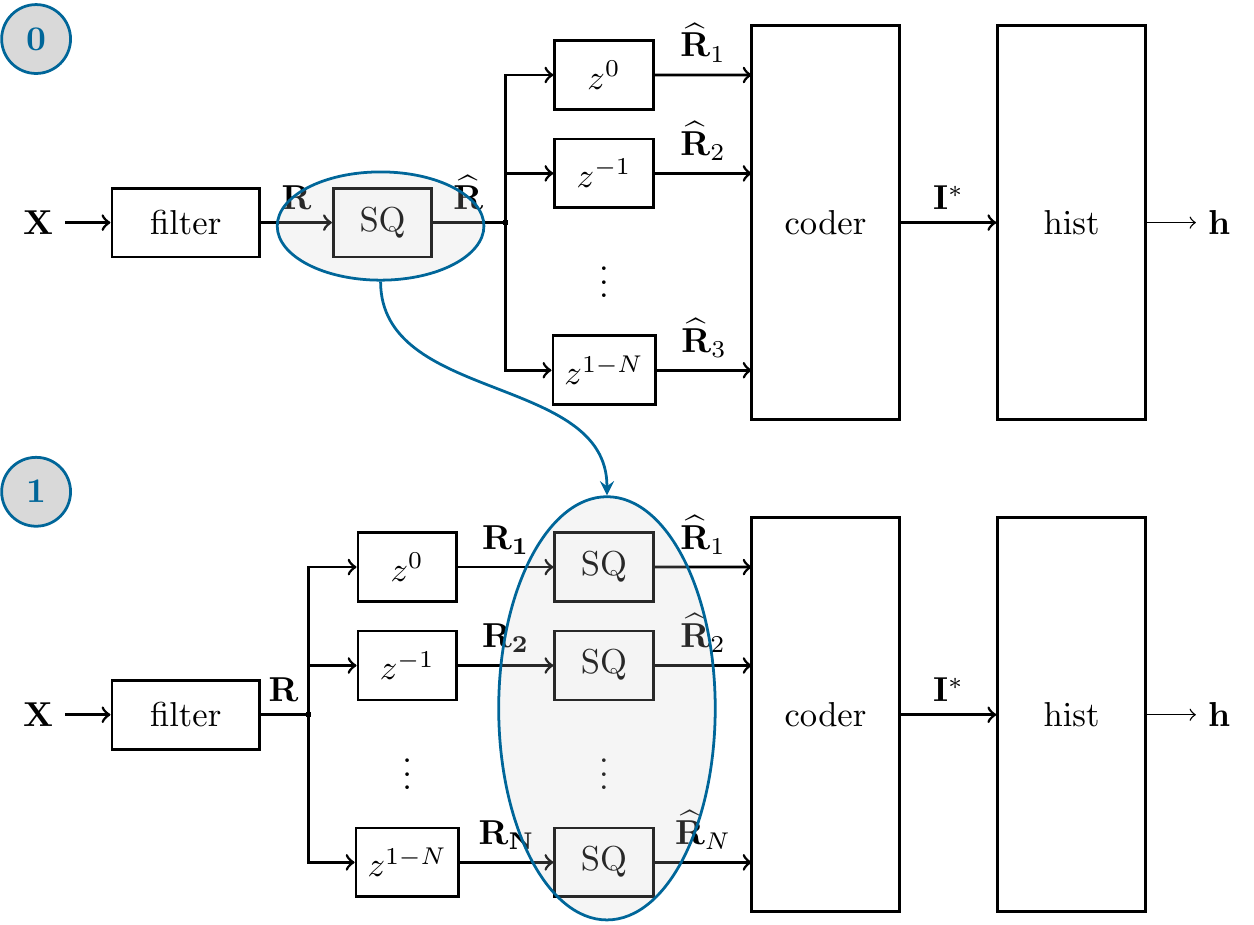}
    \caption{Basic processing scheme (0) for extracting the single model SRM feature,
    and equivalent scheme (1) with inverted order of scalar quantization (SQ) and $n$-pixel shifting ($z^{-n}$).}
    \label{schemaFridrichLD}
\end{figure}

\subsection{From local features to Bag-of-Words}

In Fig.1 (top) we show the basic processing scheme used to extract the single model SRM feature.
Let $\X$ be the input image\footnote{We use capital boldface for images, lowercase boldface for vectors, and simple lowercase for scalars.
The value of image $\X$ at spatial site $s$, will be denoted by $x_s$.}, $\R$ the residual image, and $\hR$ the quantized residual image.
To compute the output feature, the input image is high-pass filtered,
then the residual image is quantized, and $N$ versions of it are generated, shifted one pixel apart from one another.
For each pixel $s$, the values $\hr_{1,s},\ldots,\hr_{N,s}$ are regarded as base-$L$ digits and encoded as a single scalar $i^*_s$,
finally, the histogram ${\bf h}$ of this latter image is computed.
The scheme at the bottom of Fig.1 is identical to the former except for the inverted order of scalar quantization (SQ) and shifting.
This inversion, however, allows us to focus on the two groups of blocks highlighted at the top of Fig.2.

The filter-shifter group can be replaced by a bank of $N$ filters, all identical to one another except for the position of the non-zero weights.
So, with reference to our running example,
the $n$-th filter will have non-zero weights, [1, -3, 3, -1], only on the $n$-th row, and zero weights everywhere else.
Turning to the second group,
the combination of $N$ scalar quantizers can be regarded as a constrained form of vector quantization (VQ).
More specifically, it is a product VQ, since the VQ codebook is obtained as the cartesian product of the $N$ SQ codebooks.
On one hand, product quantization is much simpler and faster than VQ.
On the other hand, its strong constraints are potentially detrimental for performance.
Its $K=L^N$ codewords are forced to lie on a truncated $N$-dimensional square lattice \cite{Gray1998} and cannot adapt to the data distribution.
Many of them will be wasted in empty regions of the feature space,
causing a sure, and often severe, loss of performance with respect to unconstrained VQ.

However, the most interesting observation about the new structure at the bottom of Fig.2
is that it implements the Bag-of-Words (or also Bag-of-Features) paradigm.
The filter bank extracts a feature vector for each image pixel, based on its neighborhood.
These feature are then associated, through VQ, with some template features.
Finally, the frequency of occurrence of the latter, computed in the last block, provides a synthetic descriptor of the input image.
The fact that filters and vector quantizer are largely sub-optimal impacts only on performance, not on interpretation.
Needless to say, they could be both improved through supervised training.

\begin{figure}[t!]
	\centering
	\includegraphics[width=1.0\linewidth]{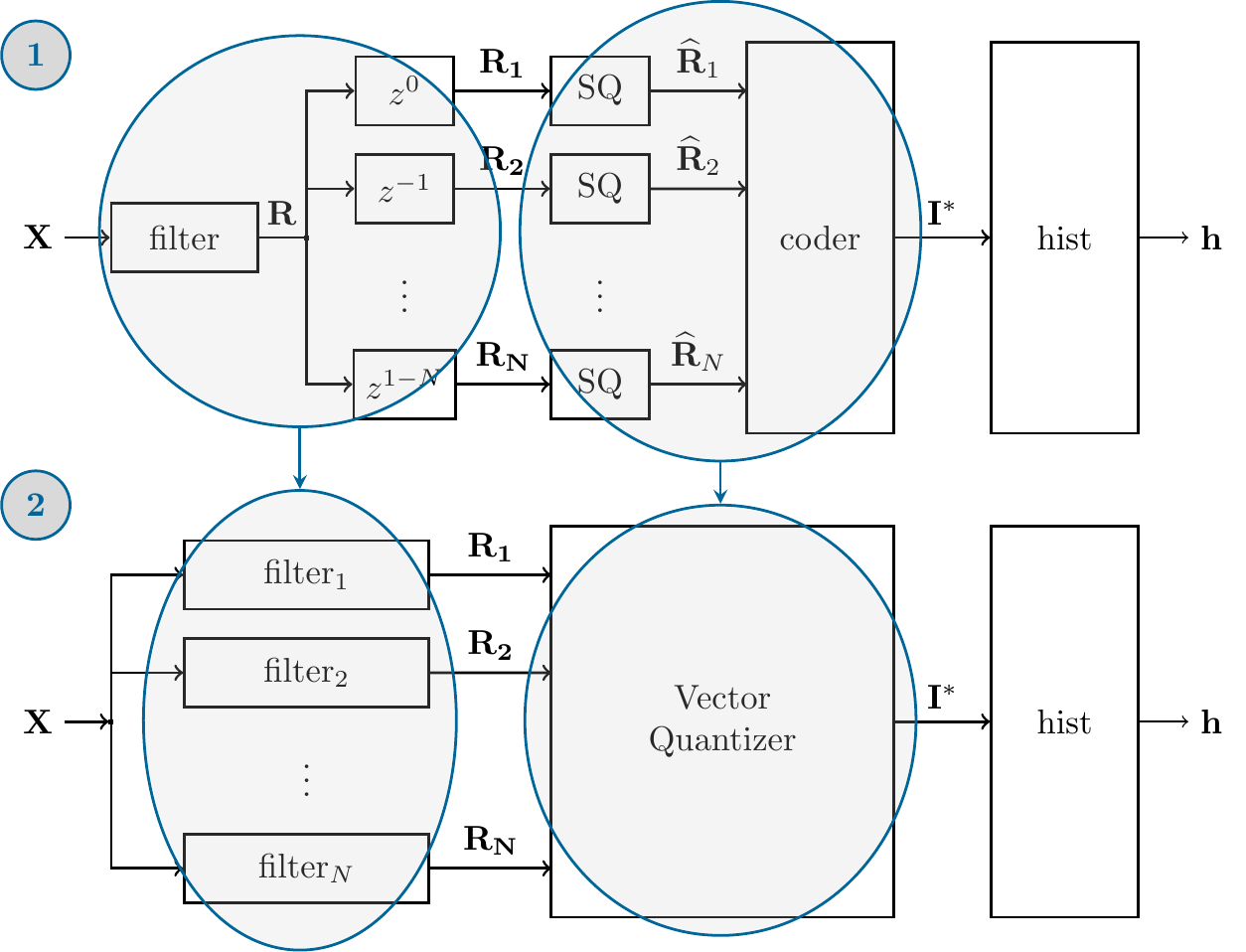}
	\caption{The cascade of filter and shifters of scheme (1) can be replaced by a bank of filters,
    while the bank of independent SQ's + coder can be replaced by product VQ.
    The resulting scheme (2) fits the Bag-of-Words paradigm.}
	\label{schemaFridrichLD}
\end{figure}

\begin{figure}[t]
	\centering
	\includegraphics[width=0.9\linewidth]{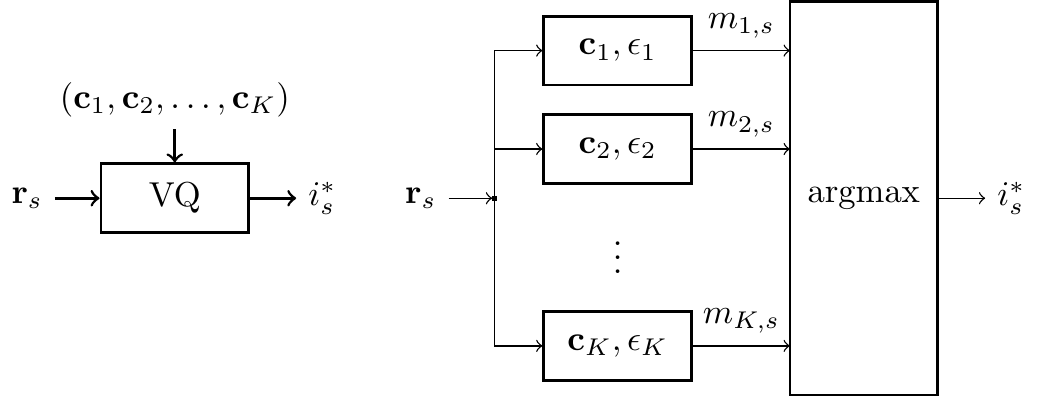}
	\caption{A vector quantizer (left) can be implemented through a bank of filter followed by argmax (right).}
	\label{schemaLD}
\end{figure}

\subsection{From Bag-of-Words to CNN}

We now show that the processing steps of Fig.2 can be all implemented through a CNN.
First of all, the bank of linear filters used to extract noise residuals can be replaced by a pure convolutional layer,
with neurons computing the residuals as
\begin{equation}
    r_{n,s} = f(w_{n,s} * x_s + b_n)   \hspace{5mm} n=1,\cdots,N
\end{equation}
with $s$ used for image spatial location and $n$ to identify neurons.
The neuron weights coincide with filter coefficients, biases $b_n$ are all set to zero, and the non-linearity $f(\cdot)$ is set to identity.

As for the vector quantizer,
assuming the usual minimum distance hard-decision rule, it can be implemented by means of a convolutional layer followed by an hard-max layer.
Let $\r_s$ be a vector formed by collecting a group of residuals at site $s$, and $\ci_k$ the $k$-th codeword of the quantizer.
Their squared Euclidean distance, $d^2_{k,s}$, can be expanded as
\begin{eqnarray}
    ||\r_s-\ci_k||^2  & = & ||\r_s||^2 + ||\ci_k||^2  - 2<\r_s,\ci_k> \nonumber\\
                      & = & ||\r_s||^2 - 2 \epsilon_k - 2<\r_s,\ci_k>
\end{eqnarray}
with $||\cdot||^2$ and $<\cdot,\cdot>$ indicating norm and inner product, respectively.
Hence, neglecting the irrelevant $||\r_s||^2$ term:
\begin{eqnarray}
    i^*_s & = & {\rm argmin}_{k=1,\cdots,K} \,\, d_{k,s} \nonumber\\
          & = & {\rm argmax}_{k=1,\cdots,K} \,\,(<\r_s,\ci_k> + \epsilon_k) \nonumber\\
          & = & {\rm argmax}_{k=1,\cdots,K} \,\, m_{k,s}
\end{eqnarray}
with $m_{k,s}$ interpreted as a matching score between the feature vector at site $s$ and the $k$-th codeword.

\begin{figure}[t!]
	\centering
	\includegraphics[width=1\linewidth]{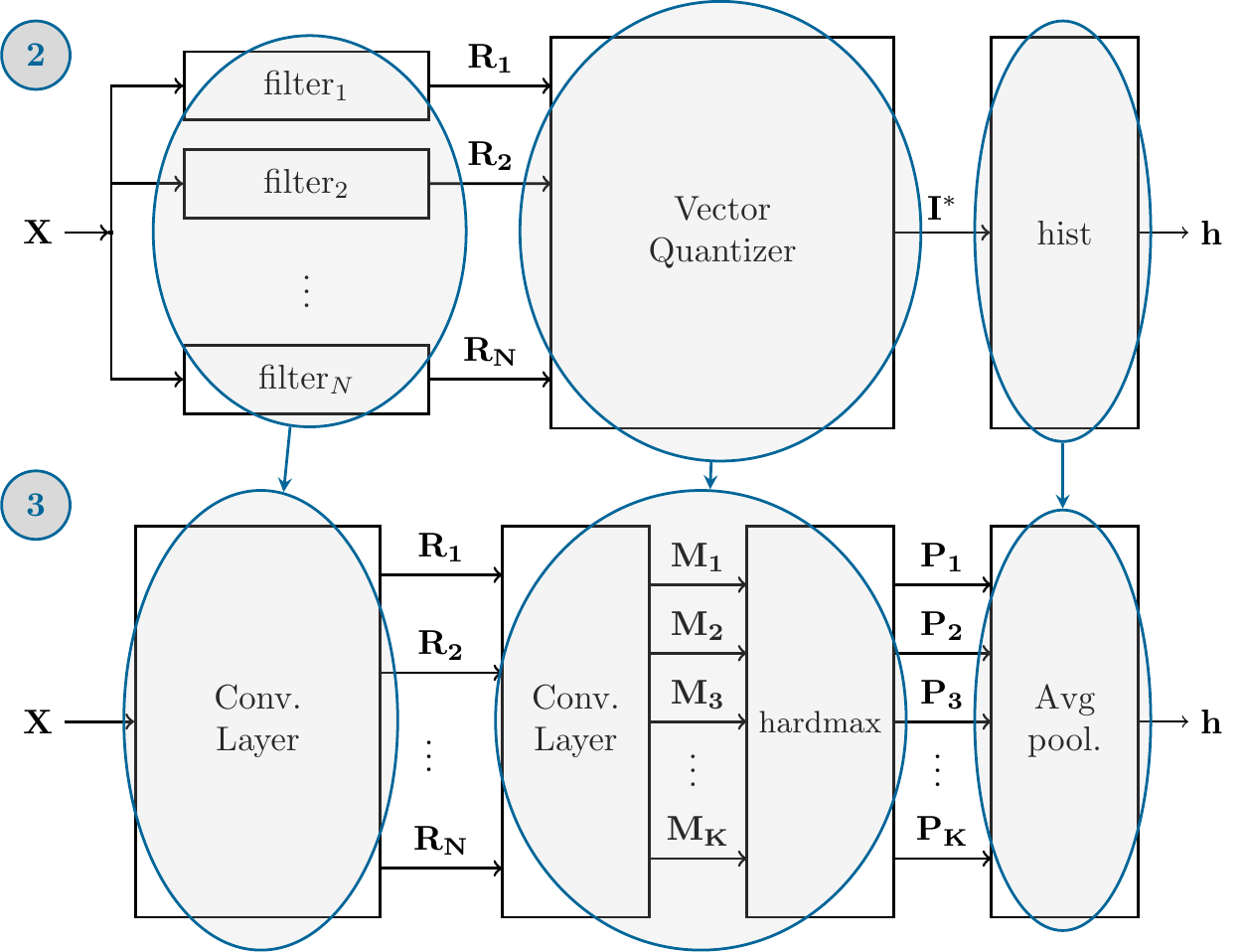}
	\caption{The whole scheme (2) can be converted in the CNN (3).
    The filter bank is replaced by a convolutional layer, VQ is replaced by convolutional-hardmax layers, the histogram can be computed through an average pooling layer.}
	\label{schemaLD}
\end{figure}

This equivalence is depicted in Fig.3.
The matching scores $m_{k,s}$ are computed through a convolutional layer,
equipped with $K=L^N$ filters (remember that $L$ is the number of quantization levels),
one for each codeword, having weights $\ci_k$, bias $\bm{\epsilon}_k$ and,
again, an identity as activation function.
The best matching codeword is then selected through a hardmax processing.

The evolution of the whole network is shown in Fig.4.
As already said, the first filter bank is replaced by a convolutional layer.
Then, the VQ is replaced by another convolutional layer followed by a hardmax layer.
The former outputs $K$ feature maps, ${\bf M}_k$, with the matching scores.
The latter outputs $K$ more binary maps, ${\bf P}_k$,
where $p_{k,s}=1$ when the corresponding matching score $m_{k,s}$ is maximum over $k$, and 0 otherwise.
Finally, the histogram computation is replaced by an average pooling layer operating on the whole feature maps, that is, $h_k = \sum_s p_{k,s}$.
The resulting scheme is shown at the bottom of Fig.4.

\begin{figure}
	\centering
	\includegraphics[width=1\linewidth]{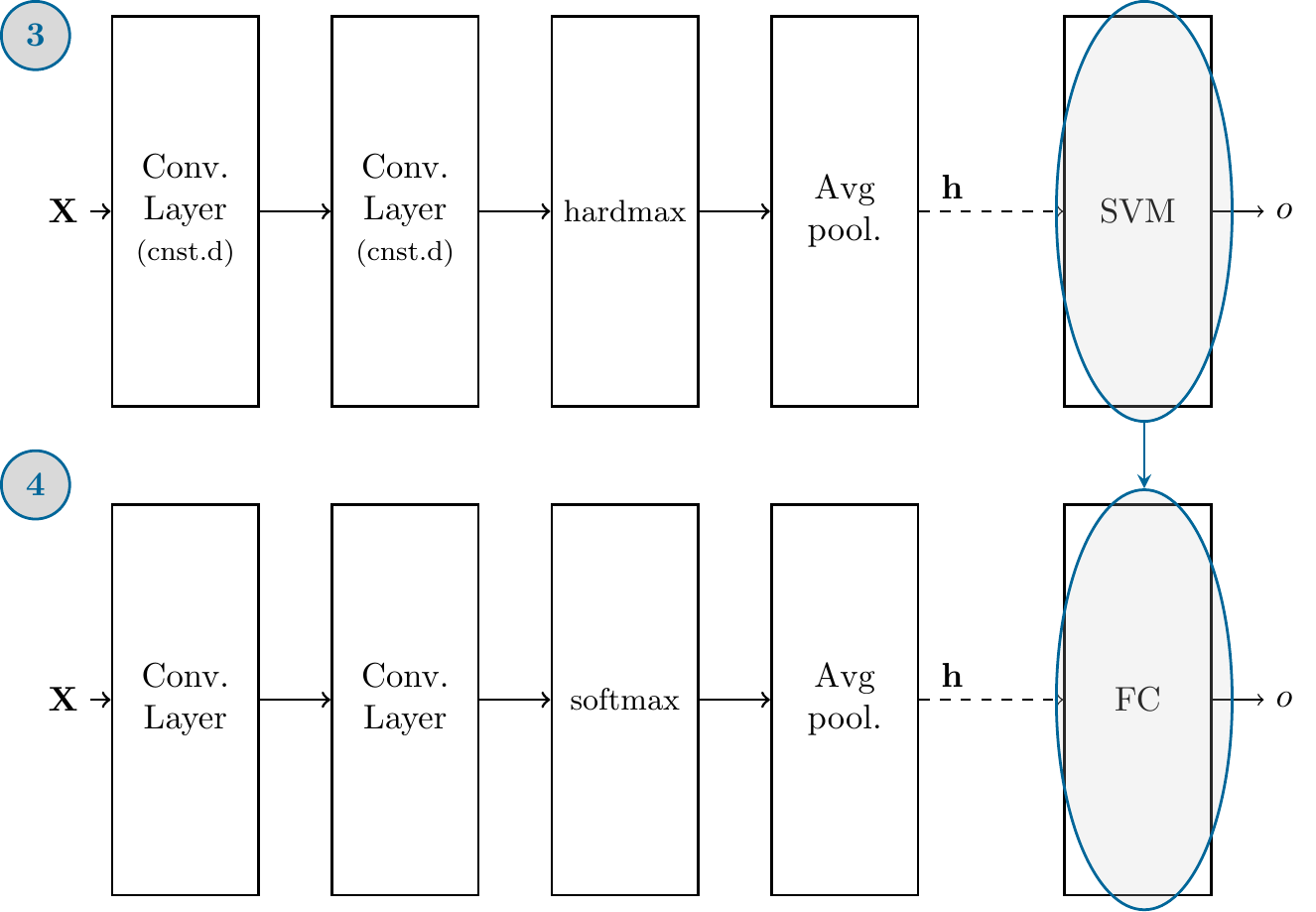}
	\caption{The constrained CNN of scheme (3) extracts the features which feed an external classifier.
    In scheme (4) this is replaced by an internal fully connected layer,
    and all constraints are removed.
    By fine-tuning on training data, all layers can be optimized jointly.
    The dashed lines are to remind that this net provides only half the feature, a twin net (not shown for clarity) provides the other half.}
	\label{schemaCnnLD}
\end{figure}

This net, actually, computes only half the desired feature, the part based on across-filter co-occurrences.
The twin half is computed similarly, and the complete feature is eventually fed to the SVM classifier, as shown in Fig.5 (top).
In this network, weights and biases of the convolutional layers
are all hard-wired to reproduce exactly the behavior of the residual-based local feature described in Section 2,
thereby ensuring the good performance observed in the literature.
We now proceed to remove all constraints and allow the net to learn on a suitable training set.
First of all, the classifier itself can be implemented as part of the CNN architecture by including a fully connected layer at the end,
obtaining the architecture of Fig.5 (bottom).
Now, to exploit the full potential of deep learning, all parameters must be optimized by appropriate training,
thereby overcoming all the impairing constraints mentioned before.
Note that the learning phase allows us not only to optimize all layers, which could be done also in the BoW framework,
but to optimize them {\em jointly}, taking full advantage of the CNN structural freedom.
Moreover, the lightweight architecture of the network
is instrumental to achieve good results even in the presence of a limited training set.

ayer with a sft-max layer that approximates it, so as to avoid non-differentiable operators.

However, before proceeding with the training, it is necessary to replace the hard-max layer,
with a soft-max layer that approximates it, so as to avoid non-differentiable operators.
Given the input vector $\{m_{k,s}, k=1,\ldots,K\}$, the soft-max computes the quantities
\begin{equation}
    p_{k,s} = e^{\alpha m_{k,s}}/\sum_l e^{\alpha m_{l,s}}
\end{equation}
With the aim to preserve a close correspondence with the original descriptor, we should choose a very large $\alpha$ parameter, so as to obtain a steep nonlinearity.
However, as said before, this is not really necessary, since our goal is only to improve performance.
Hence, we select a relatively small value for $\alpha$ in order not to slow down learning.
Likewise, to implement minimum distance VQ exactly, the biases in the second convolutional layer should depend on the filter weights,
but there is no practical reason to enforce this constraint, and we allow also the biases to adapt freely.

Now, the final CNN can be trained as usual with stochastic gradient descent \cite{Krizhevsky2012} to adapt to the desired task.
It should be clear that a number of architectural modifications could be also tested starting from this basic structure,
but this goes beyond the scope of the present paper, and will be the object of future research.
We must underline that the equivalence between CNN and BoW has been noticed before in the literature,
for example in \cite{Lan2015, Arandjelovic2016, Richard2016}.

\section{Experimental analysis}

\renewcommand{\tabcolsep}{6pt}
\begin{table}
    \caption{Image manipulations under test.}
	\begin{center}
		\begin{tabular}{|l|l|} \hline
			\ru			Manipulation           & Parameters                                 \\ \hline \hline
			\ru			Median Filtering       & kernel: 7$\times$7, 5$\times$5, 3$\times$3 \\ \hline
			\ru			Gaussian Blurring      & st. dev: 1.1,\; 0.75,\; 0.5                \\ \hline
			\ru			Additive Noise  (AWGN) & st. dev.: 2.0,\; 0.5,\; 0.25               \\ \hline
			\ru			Resizing               & scale: 1.5,\; 1.125,\; 1.01                \\ \hline
			\ru			JPEG Compression       & quality factor: 70,\;  80,\; 90            \\ \hline
		\end{tabular}
	\end{center}
	\label{tab:type}
\end{table}

\renewcommand{\tabcolsep}{6pt} \begin{table*}
	\label{tab:perf}
	\caption{Detection Accuracy for binary classification tasks.}
	\begin{center}
		\begin{tabular}{|c|c||c|c|c||c|c|c|}  \hline
			\multicolumn{2}{|c||}{\ru Manipulation} & \multicolumn{3}{c||}{Small Training Set} & \multicolumn{3}{c|}{Large Training Set} \\ \hline
			\multicolumn{2}{|c||}{\ru}              &   Bayar2016 &     SRM+SVM &    prop. CNN &   Bayar2016 &     SRM+SVM &   prop. CNN \\
			\multicolumn{2}{|c||}{   }              &   60 epochs &             &    15 epochs &   60 epochs &             &   15 epochs \\ \hline \hline
			                  & \ru    7x7          &       98.23 &       99.61 &       99.07  &       99.69 &       99.68 &       99.55 \\ \cline{2-8}
			Median Filtering  & \ru    5x5          &       96.66 &       99.67 &       99.47  &       99.78 &       99.68 &       99.60 \\ \cline{2-8}
			                  & \ru    3x3          &       94.56 &       99.83 &       99.35  &       99.80 &       99.87 &       99.75 \\ \hline \hline
			                  & \ru    1.1          &       99.65 &       99.93 &       99.79  &       99.98 &       99.97 &       99.95 \\ \cline{2-8}
			Gaussian Blurring & \ru    0.75         &       98.52 &       99.90 &       99.82  &       99.94 &       99.77 &       99.93 \\ \cline{2-8}
			                  & \ru    0.5          &       83.10 &       87.10 &       95.70  &       94.57 &       87.55 &       96.56 \\ \hline \hline
			                  & \ru    2.0          &       97.08 &       99.94 &       99.95  &       99.56 &       99.94 &       99.94 \\ \cline{2-8}
			Additive Noise    & \ru    0.5          &       82.93 &       99.37 &       99.36  &       93.83 &       99.34 &       99.66 \\ \cline{2-8}
			                  & \ru    0.25         &       51.83 &       85.06 &       88.81  &       80.28 &       84.01 &       90.79 \\ \hline \hline
			                  & \ru    1.5          &       99.22 &       99.99 &      100.00  &       99.72 &       99.87 &      100.00 \\ \cline{2-8}
			Resizing          & \ru ~~1.125~~       &       91.06 &       98.94 &       99.56  &       97.02 &       96.00 &       99.78 \\ \cline{2-8}
			                  & \ru    1.01         &       80.51 &       96.01 &       97.81  &       98.44 &       95.11 &       97.20 \\ \hline \hline
			                  & \ru     70          &       96.04 &       99.99 &       99.99  &       99.43 &       99.99 &       99.94 \\ \cline{2-8}
			JPEG Compression  & \ru     80          &       77.01 &       99.73 &       99.37  &       98.12 &       99.94 &       99.86 \\ \cline{2-8}
			                  & \ru     90          &       63.77 &       90.86 &       92.08  &       79.69 &       90.90 &       94.59  \\ \hline
		\end{tabular}
	\end{center}
\end{table*}

To test the performance of the proposed CNN architecture we carry out a number of experiments with typical manipulations.
Our synthetic dataset includes images taken from 9 devices,
4 smartphones (Apple iPhone 4S, Apple iPhone 5s, Huawei P7 mini, Nokia Lumia 925) and 5 cameras (Canon EOS 450D, Canon IXUS 95 IS, Sony DSC-S780, Samsung Digimax 301, Nikon Coolpix S5100).
Each device contributes 200 images, and from each image non-overlapping patches of dimension $128\times 128$ are sampled.
We select at random 6 devices to form the training set, while the remaining 3 are used as a testing set.
Therefore, the patches used for testing come from devices that are never seen in the training phase.
For each pristine patch, the corresponding manipulated patch is also included in the set.
Overall, our training set comprises a total of just 10800+10800 patches, quite a small number for deep learning applications.
We consider 5 types of image manipulation: median filtering, gaussian blurring, AWGN noise addition, resizing, and JPEG compression,
with three different settings for each case (see Tab.1) corresponding to increasingly challenging tasks.
For example, JPEG compression with quality factor Q=70 is always easily detected,
while a quality factor Q=90 makes things much more difficult.

In the proposed CNN
the first convolutional layer includes 4 filters of size $5\times5\times1$ operating on the monochrome input (we use only the green band normalized in $[0,1]$.
In the second layer there are 81 filters of size $1\times1\times4$.
Filters are initialized as described in Section 3 and the $\alpha$ parameter of soft-max is set to $2^{16}$.
The code is implemented in Tensorfow and runs on a Nvidia Tesla P100 with 16GB RAM.
We set the learning rate to $10^{-6}$, with decay $5 \cdot 10^{-4}$, batch size 36 and Adam \cite{Kingma2015} optimization method, using the cross-entropy loss function.
Together with the proposed CNN we consider also the basic solution,
with the handcrafted feature followed by linear SVM,
and the CNN proposed in Bayar2016 \cite{Bayar2016}
based on the use of a preliminary high-pass convolutional layer.

Results in terms of probability of correct decision for each binary classification problem
are reported in the left part of Tab.2 (small training set).
With ``easy'' manipulations, {\it e.g.} JPEG@70, all methods provide near-perfect results
and there is no point in replacing the SRM+SVM solution with something else.
In the presence of more challenging attacks, however, the performance varies significantly across methods.
After just 15 epochs of fine tuning, the proposed CNN improves over SRM+SVM of about 2 percent points for JPEG compression, resizing, and noising, and more than 8 points for blurring,
while median filtering is almost always detected in any case.
In the same cases, the CNN architecture proposed in \cite{Bayar2016} provides worse results,
sometimes close to 50\%, even after 60 epochs of training.
Our conjecture is that a deep CNN is simply not able to adapt correctly with a small training set.
In this condition a good hand-crafted feature can work much better.
The proposed CNN builds upon this result and takes advantage of the available limited training data
to fine-tune its parameters.

To carry out a fair comparison we also considered a case in which a much larger training set is available,
comprising 460800 patches, that is more than 20 times larger than before.
Results are reported in the right part of Tab.2.
As expected the performance does not change much for the SRM+SVM solution,
since the SVM needs limited training anyway.
For the proposed CNN, some improvements are observed for the more challenging tasks.
As an example, for JPEG@90 the accuracy grows from 92.08 to 94.59.
Much larger improvements are observed for the network proposed in  \cite{Bayar2016},
which closes almost always the performance gap and sometimes outperforms slightly the proposed CNN.
Nonetheless in a few challenging cases,
like the already mentioned JPEG@90 or the addition of low-power white noise,
there is still a difference of more than 10 percent point with our proposal.
It is also interesting to compare the two adjacent columns,
proposed CNN at 15 epochs and the CNN architecture proposed in \cite{Bayar2016} at 60 epochs,
which speak clearly in favor of the first solution, in terms of both complexity and performance.


\begin{figure*}[t!]
	\centering
	\begin{tabular}{cccc}
		\includegraphics[width=0.22\linewidth]{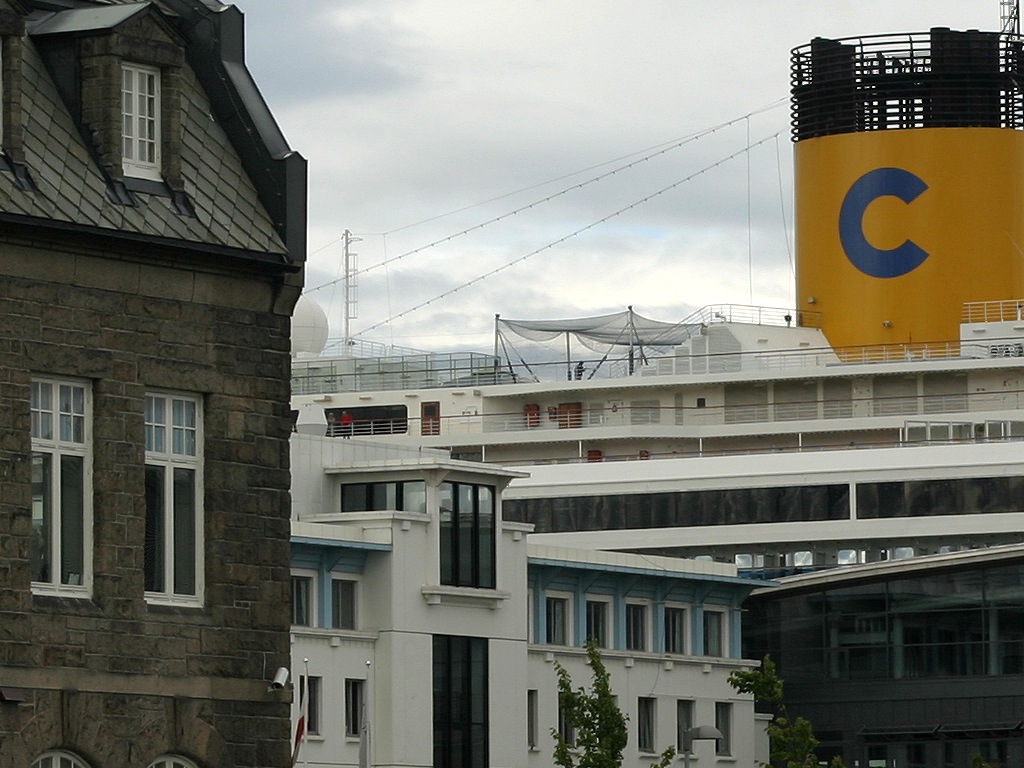} \;
		\includegraphics[width=0.22\linewidth]{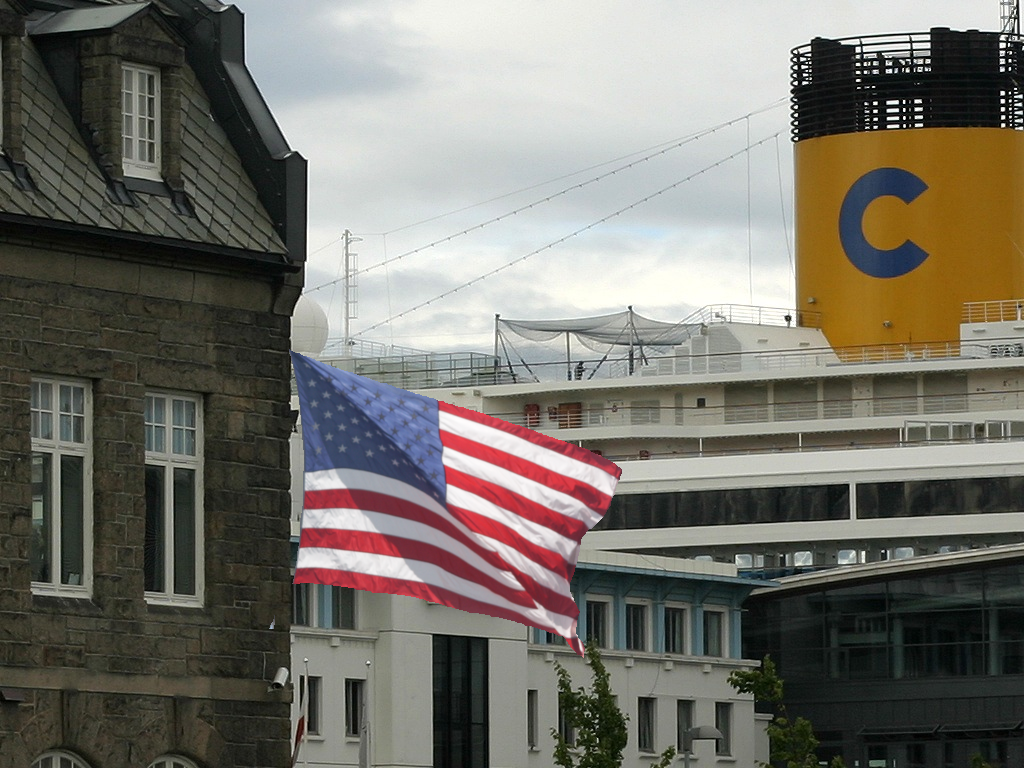} \;
		\includegraphics[width=0.22\linewidth]{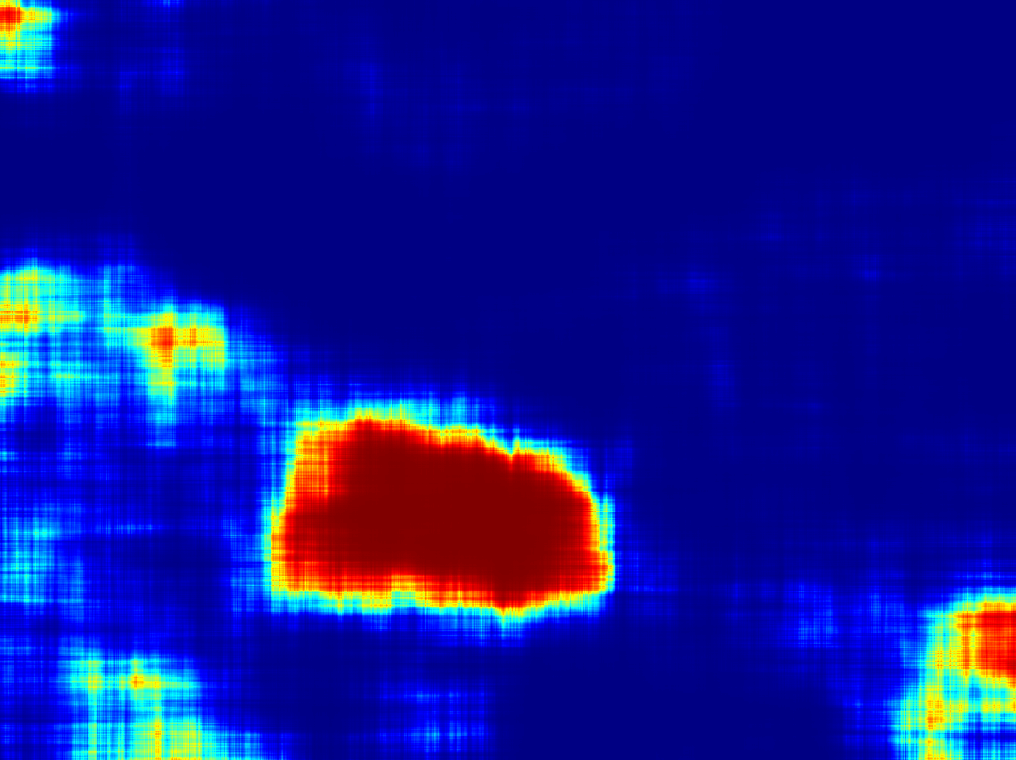} \;
		\includegraphics[width=0.22\linewidth]{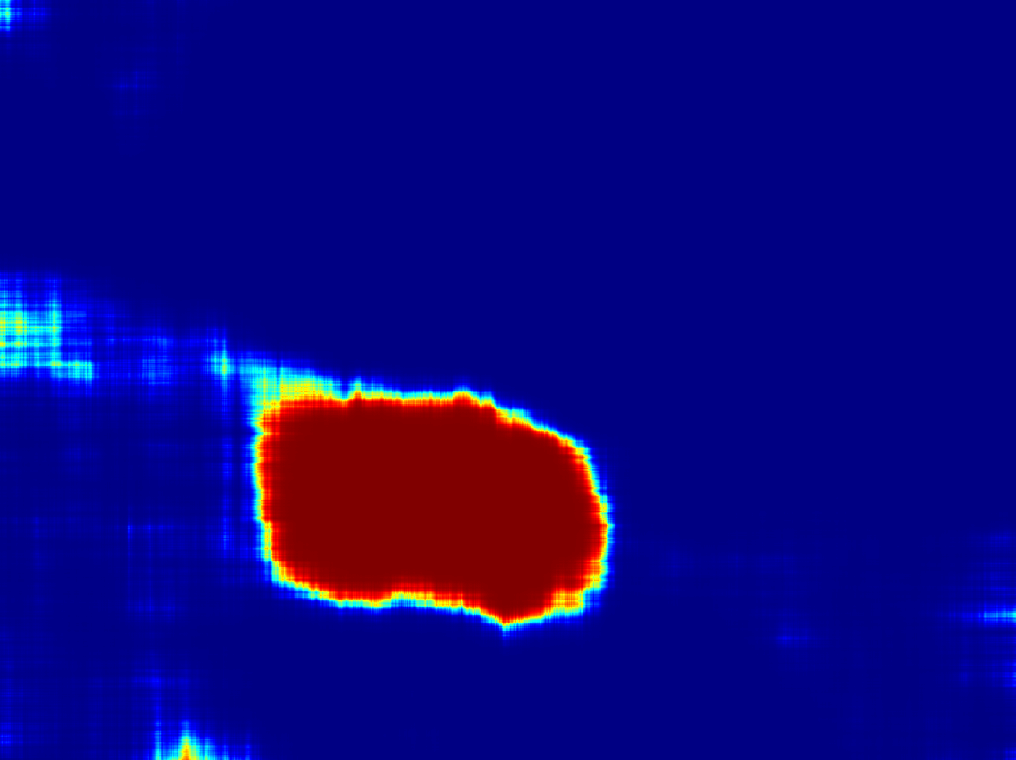}  \\ \vspace{-3mm} \\
		\includegraphics[width=0.22\linewidth]{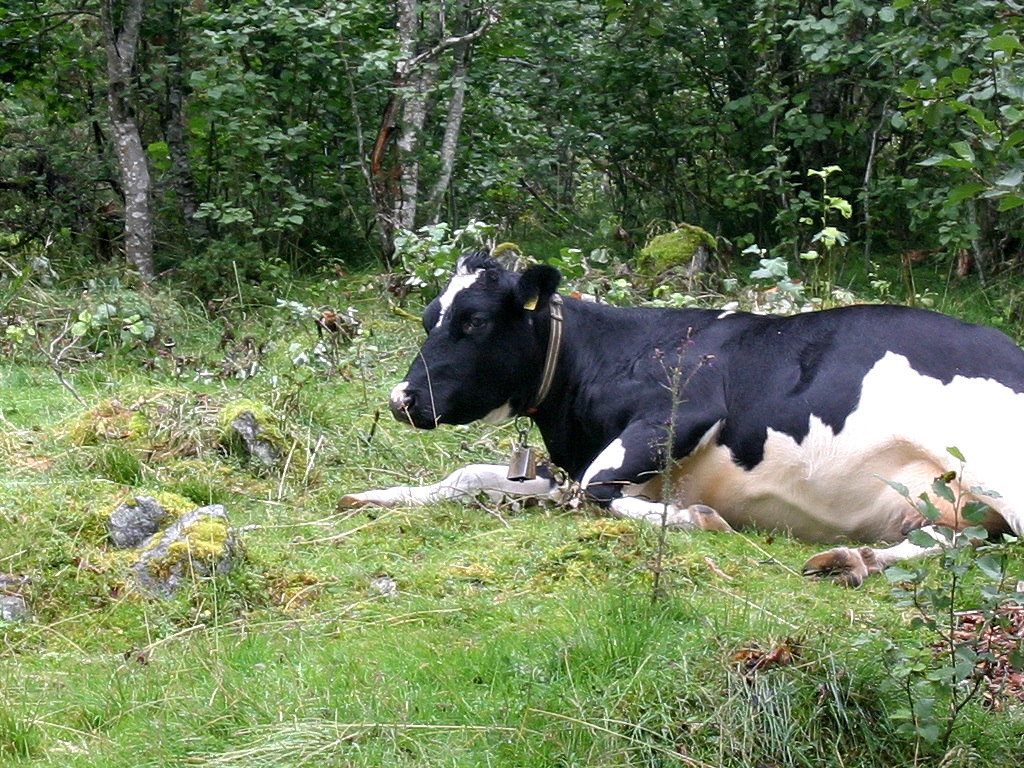} \;
		\includegraphics[width=0.22\linewidth]{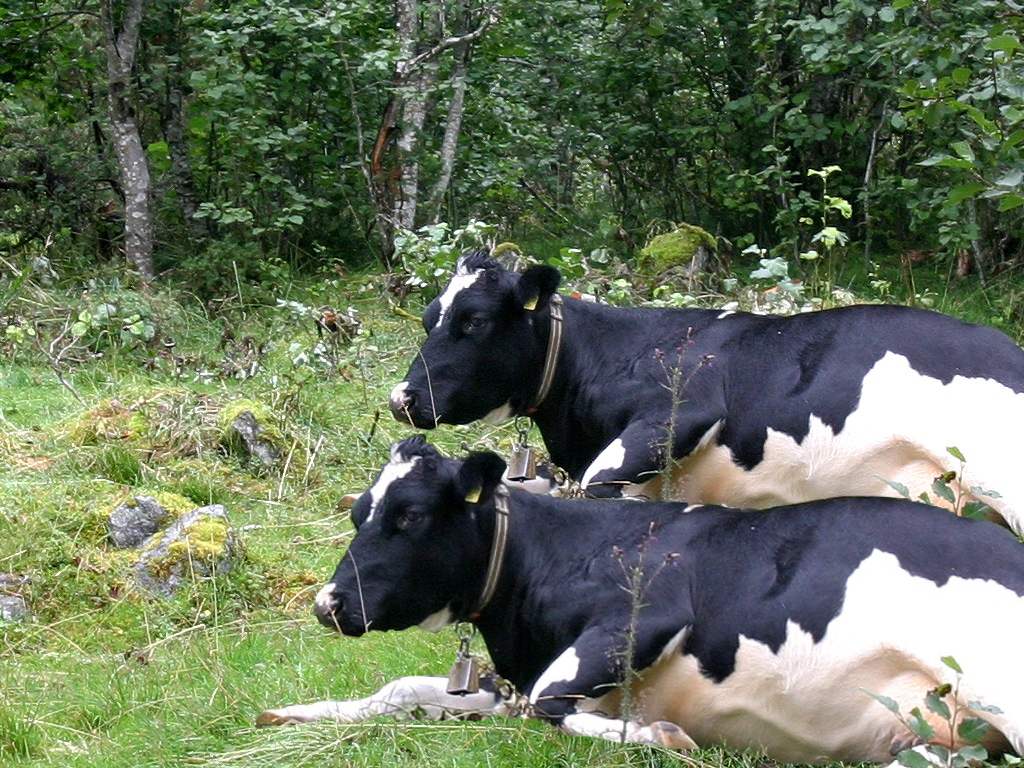} \;
		\includegraphics[width=0.22\linewidth]{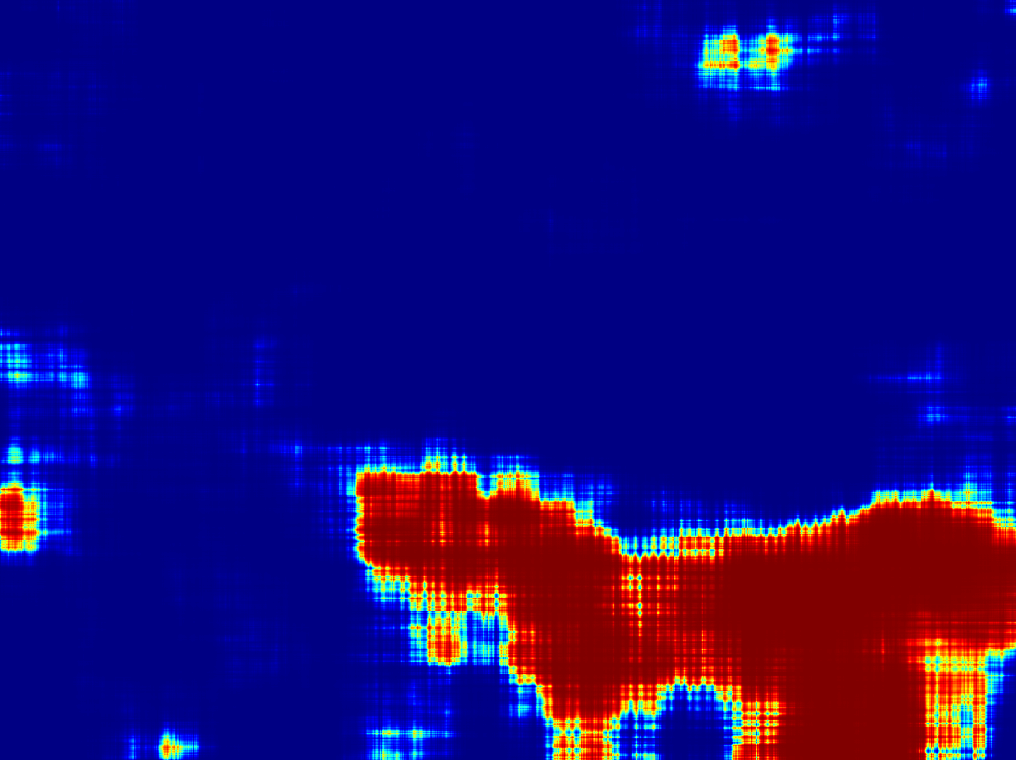} \;
		\includegraphics[width=0.22\linewidth]{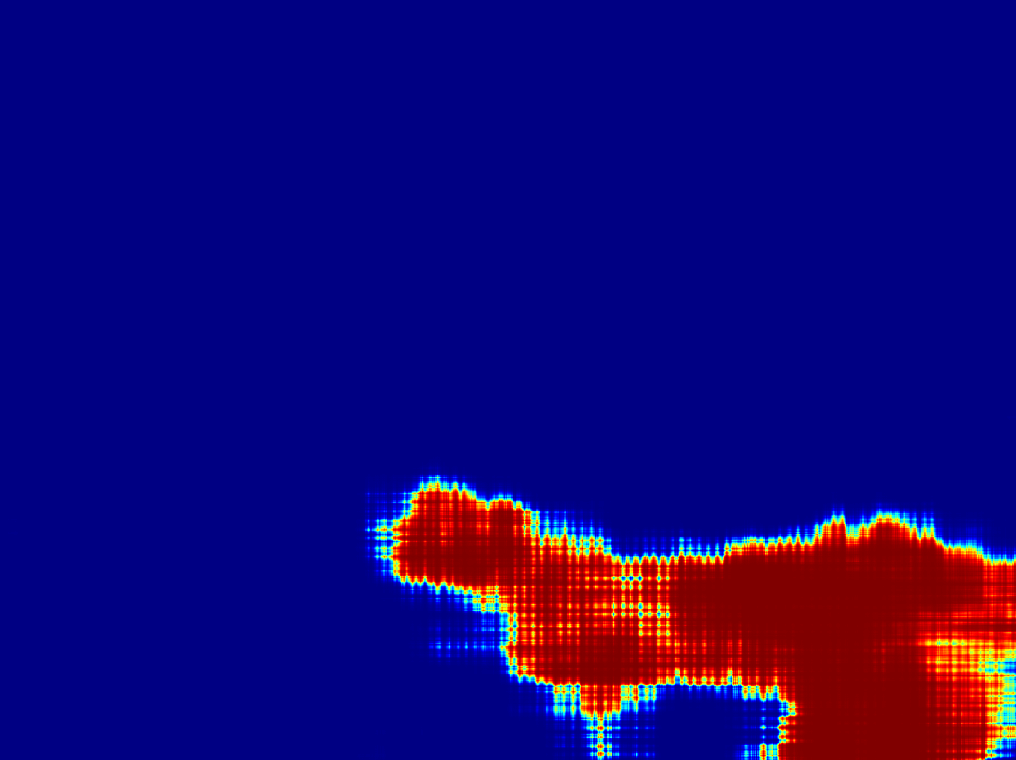}
	\end{tabular}	
	\caption{Top: splicing (blurred with st. dev. 0.5). Bottom: copy-move (resized with scale 1.125).
		From left to right: original image, forged image, SRM+SVM heat map, proposed CNN heat map.}
	\label{fig:localization1}
\end{figure*}

The ability to reliably classify small
patches may be very valuable in the presence of spatially localized attacks.
This is the case of image copy-move or splicing, where only a small part of the image is tampered with.
In these cases, a descriptor computed on small patches can more reliably detect manipulations,
and even localize the forgery by working in sliding-window modality.
Fig.6 shows two examples of forgery localization with a slightly blurred splicing and a resized copy-move, respectively.
In both cases
using the proposed CNN in sliding-window modality, a sharp heat map is obtained,
allowing for a precise localization of the forgery.
The SRM+SVM solution also provides good results, but the heat map is more fuzzy, with a higher risk of false alarms.
It is worth underlining again that the images used for these tests come from cameras that did not contribute to the training set.

\section{Conclusions}

Residual-based descriptors have proven extremely effective for a number of image forensic applications.
Improving upon the current state of the art, however, is slow and costly,
since the design of better hand-crafted features is not trivial.
We showed that a class of residual-based features can be regarded as compact constrained CNNs.
This represent a precious starting point to exploit the huge potential of deep learning,
as testified by the promising early results.
However, this is only a first step,
and there is much room for improvements, especially through new architectural solutions.
This will be the main focus of future work.

\balance


\vspace{3mm}
{\footnotesize
{\bf Acknowledgment.}
This material is based on research \vspace{-0.6mm}
sponsored by the Air Force Research Laboratory \vspace{-0.6mm}
and the Defense Advanced Research Projects Agency \vspace{-0.6mm}
under agreement number FA8750-16-1-0104.
The U.S. Government is authorized to reproduce and distribute reprints \vspace{-0.6mm}
for Governmental purposes notwithstanding any copyright notation thereon. \vspace{-0.6mm}
The views and conclusions contained herein are those of the authors \vspace{-0.6mm}
and should not be interpreted as necessarily representing the official policies or endorsements, \vspace{-0.6mm}
either expressed or implied, of the Air Force Research Laboratory \vspace{-0.6mm}
and the Defense Advanced Research Projects Agency or the U.S. Government.
}

\bibliographystyle{IEEEtran}
\bibliography{forensic}

\begin{thebibliography}{10}
\providecommand{\url}[1]{#1}
\csname url@samestyle\endcsname
\providecommand{\newblock}{\relax}
\providecommand{\bibinfo}[2]{#2}
\providecommand{\BIBentrySTDinterwordspacing}{\spaceskip=0pt\relax}
\providecommand{\BIBentryALTinterwordstretchfactor}{4}
\providecommand{\BIBentryALTinterwordspacing}{\spaceskip=\fontdimen2\font plus
\BIBentryALTinterwordstretchfactor\fontdimen3\font minus
  \fontdimen4\font\relax}
\providecommand{\BIBforeignlanguage}[2]{{%
\expandafter\ifx\csname l@#1\endcsname\relax
\typeout{** WARNING: IEEEtran.bst: No hyphenation pattern has been}%
\typeout{** loaded for the language `#1'. Using the pattern for}%
\typeout{** the default language instead.}%
\else
\language=\csname l@#1\endcsname
\fi
#2}}
\providecommand{\BIBdecl}{\relax}
\BIBdecl

\bibitem{Popescu2005}
A.~Popescu and H.~Farid, ``Exposing digital forgeries by detecting traces of
  resampling,'' \emph{IEEE Transactions on Signal Processing}, vol.~53, no.~2,
  pp. 758--757, 2005.

\bibitem{Kirchner2008}
M.~Kirchner, ``Fast and reliable resampling detection by spectral analysis of
  fixed linear predictor residue,'' in \emph{Proceedings of the Multimedia and
  Security Workshop}, 2008, pp. 11--20.

\bibitem{Kirchner2010}
M.~Kirchner and J.~Fridrich, ``On detection of median filtering in digital
  images,'' in \emph{SPIE, Electronic Imaging, Media Forensics and Security
  XII}, vol. 7541, 2010, pp. 101--112.

\bibitem{Huang2010}
F.~Huang, J.~Huang, and Y.~Shi, ``{Detecting double JPEG compression with the
  same quantization matrix},'' \emph{IEEE Transactions on Information Forensics
  and Security}, vol.~5, no.~4, pp. 848--856, dec 2010.

\bibitem{Stamm2010}
M.~Stamm and K.~R. Liu, ``Forensic detection of image manipulation using
  statistical intrinsic fingerprints,'' \emph{IEEE Transactions on Information
  Forensics and Security}, vol.~5, no.~3, pp. 492--506, september 2010.

\bibitem{Cao2012}
H.~Cao and A.~Kot, ``Manipulation detection on image patches using
  fusionboost,'' \emph{IEEE Transactions on Information Forensics and
  Security}, vol.~7, no.~3, pp. 992--1002, june 2012.

\bibitem{Verdoliva2014}
L.~Verdoliva, D.~Cozzolino, and G.~Poggi, ``A feature-based approach for image
  tampering detection and localization,'' in \emph{IEEE Workshop on Information
  Forensics and Security}, 2014, pp. 149--154.

\bibitem{Fan2015}
W.~Fan, K.~Wang, and F.~Cayre, ``General-purpose image forensics using patch
  likelihood under image statistical models,'' in \emph{IEEE International
  Workshop on Information Forensics and Security}, 2015, pp. 1--6.

\bibitem{Li2016}
H.~Li, W.~Luo, X.~Qiu, and J.~Huang, ``Identification of various image
  operations using residual-based features,'' \emph{IEEE Transactions on
  Circuits and Systems for Video Technology, in press}, 2016.

\bibitem{Cozzolino2014a}
D.~Cozzolino, D.~Gragnaniello, and L.~Verdoliva, ``Image forgery detection
  through residual-based local descriptors and block-matching,'' in \emph{IEEE
  International Conference on Image Processing}, 2014, pp. 5297--5301.

\bibitem{Cozzolino2014b}
------, ``Image forgery localization through the fusion of camera-based,
  feature-based and pixel-based techniques,'' in \emph{IEEE International
  Conference on Image Processing}, 2014, pp. 5302--5306.

\bibitem{Pevny2010}
T.~Pevn{\'{y}}, P.~Bas, and J.~Fridrich, ``Steganalysis by subtractive pixel
  adjacency matrix,'' \emph{IEEE Transactions on Information Forensics and
  Security}, vol.~5, no.~2, pp. 215--224, june 2010.

\bibitem{Fridrich2012}
J.~Fridrich and J.~Kodovsk{\'{y}}, ``Rich models for steganalysis of digital
  images,'' \emph{IEEE Transactions on Information Forensics and Security},
  vol.~7, no.~3, pp. 868--882, june 2012.

\bibitem{Cozzolino2015b}
D.~Cozzolino, G.~Poggi, and L.~Verdoliva, ``Splicebuster: a new blind image
  splicing detector,'' in \emph{IEEE International Workshop on Information
  Forensics and Security}, 2015, pp. 1--6.

\bibitem{Cozzolino2016}
D.~Cozzolino and L.~Verdoliva, ``Single-image splicing localization through
  autoencoder-based anomaly detection,'' in \emph{IEEE Workshop on Information
  Forensics and Security}, 2016, pp. 1--6.

\bibitem{Boroumand2017}
M.~Boroumand and J.~Fridrich, ``{Scalable Processing History Detector for JPEG
  Images},'' in \emph{IS\&T Electronic Imaging - Media Watermarking, Security,
  and Forensics}, 2017.

\bibitem{Krizhevsky2012}
A.~Krizhevsky, I.~Sutskever, and G.~E. Hinton, ``Imagenet classification with
  deep convolutional neural networks,'' in \emph{Conference on Neural
  Information Processing Systems}, 2012, pp. 1097--1105.

\bibitem{Bengio2013}
Y.~Bengio, A.~Courville, and P.~Vincent, ``{Representation Learning: A Review
  and New Perspectives},'' \emph{IEEE Transactions on Pattern Analysis and
  Machine Intelligence}, vol.~35, no.~8, pp. 1798--1828, 2013.

\bibitem{LeCun2015}
Y.~LeCun, Y.~Bengio, and G.~Hinton, ``Deep learning,'' \emph{Nature}, vol. 521,
  pp. 436--444, may 2015.

\bibitem{Zeiler2014}
M.~Zeiler and R.~Fergus, ``Visualizing and understanding convolutional
  networks,'' in \emph{European Conference on Computer Vision}, vol. 8689,
  2014, pp. 818--833.

\bibitem{Qian2015}
Y.~Qian, J.~Dong, W.~Wang, and T.~Tan, ``Deep learning for steganalysis via
  convolutional neural networks,'' in \emph{IS\&T/SPIE Electronic Imaging},
  2015, pp. 94\,090J--94\,090J.

\bibitem{Bayar2016}
B.~Bayar and M.~Stamm, ``A deep learning approach to universal image
  manipulation detection using a new convolutional layer,'' in \emph{ACM
  Workshop on Information Hiding and Multimedia Security}, 2016, pp. 5--10.

\bibitem{Gray1998}
R.~M. Gray and D.~L. Neuhoff, ``Quantization,'' \emph{IEEE Transactions on
  Information Theory}, vol.~44, no.~6, pp. 2325--2383, 1998.

\bibitem{Lan2015}
Z.~Lan, S.-I. Yu, M.~Lin, B.~Raj, and A.~Hauptmann, ``Local handcrafted
  features are convolutional neural networks,'' \emph{arXiv preprint
  arXiv:1511.05045v2}, nov 2015.

\bibitem{Arandjelovic2016}
R.~Arandjelovic, P.~Gronat, A.~Torii, T.~Pajdla, and J.~Sivic, ``{NetVLAD: CNN
  architecture for weakly supervised place recognition},'' in \emph{IEEE
  International Conference on Computer Vision}, 2016, pp. 5297--5307.

\bibitem{Richard2016}
A.~Richard and J.~Gall, ``A bag-of-words equivalent recurrent neural network
  for action recognition,'' \emph{Computer Vision and Image Understanding},
  vol. in press, 2016.

\bibitem{Kingma2015}
D.~Kingma and J.~Ba, ``Adam: A method for stochastic optimization,'' in
  \emph{International Conference on Learning Representations (ICLR)}, 2015.

\end{thebibliography}

\end{document}